# Strategic Innovation Management in the Age of Large Language Models Market Intelligence, Adaptive R&D, and Ethical Governance

Raha Aghaei[1,2], Ali A. Kiaei[2,3]*, Mahnaz Boush[4], Mahan Rofoosheh[5], Mohammad Zavvar[6]

**Highlights:**

- Large Language Models (LLMs) considerably increase strategic innovation management by automating knowledge discovery, enabling predictive analytics, and integrating transdisciplinary insights.
- LLMs empower adaptive and tailored innovation techniques through continuous learning, real-time data integration, and agile product development cycles.
- The deployment of LLMs enhances multilingual cooperation and democratizes data-driven decision-making across global innovation ecosystems.
- Ethical governance, bias mitigation, and responsible AI frameworks are necessary to assure fairness, openness, and trustworthiness in LLM-powered innovation processes.
- Future innovation management will benefit from multimodal AI integration, privacy-preserving learning methodologies, and sustainability-focused AI applications supported by transparent and explainable AI models.

## Abstract

This study analyzes the multiple functions of Large Language Models (LLMs) in transforming research and development (R&D) processes. By automating knowledge discovery, boosting hypothesis creation, integrating transdisciplinary insights, and enabling cooperation within innovation ecosystems, LLMs dramatically improve the efficiency and effectiveness of research processes. Through extensive analysis of scientific literature, patent databases, and experimental data, these models enable more flexible and informed R&D workflows, ultimately accelerating innovation cycles and lowering time-to-market for breakthrough ideas.

---

[1] School of Mathematics and Computer Science, Iran University of Science & Technology, Tehran, Iran

[2] Department of Computer engineering, Sharif University of Technology, Tehran, Iran

[3] (Correspondence: ali.kiaei@sharif.edu) Department of Artificial Intelligence in Medicine, Faculty of Advanced Technologies in Medicine, Iran University of Medical Sciences, Tehran, Iran

[4] (Correspondence: m.boush@sbmu.ac.ir) Cellular and Molecular Biology Research Center, Shahid Beheshti University of Medical Sciences, Tehran, Iran

[5] Computer engineering group, Alborz Vocational Technical University, Alborz, Iran

[6] Department of Computer Engineering Sari Branch, Islamic Azad University, Sari, Iran Department of Computer Engineering Sari Branch, Islamic Azad University, Sari, Iran

# 1- Introduction: Strategic Challenges in Innovation Management

In today's fast-evolving corporate market, innovation is no longer a simply operational function but a crucial strategic imperative that molds the competitive advantage and long-term sustainability of firms. The acceleration of technological progress, globalization of markets, and increasingly sophisticated customer expectations have transformed innovation management into a complex, multifaceted discipline requiring not only creativity and research but also robust strategic foresight and agile decision-making. [1–5]

Large Language Models (LLMs), as state-of-the-art AI systems, offer unprecedented capabilities to process and synthesize enormous volumes of unstructured data, enabling enterprises to extract actionable insights in real time. Their integration into innovation management heralds a new era where strategic decision-making is reinforced by AI-driven intelligence, permitting enhanced market sensing, predictive analytics, and adaptive strategy development. [6–10]

Large Language Models (LLMs), as state-of-the-art AI systems, offer unprecedented ability to process and synthesize massive volumes of unstructured data, enabling organizations to extract actionable insights in real time. Their integration into innovation management heralds a new era where strategic decision-making is reinforced by AI-driven intelligence, permitting increased market sensing, predictive analytics, and adaptive strategy building. [11–13]

This article addresses the strategic constraints and opportunities inherent in adopting LLMs within innovation management. It focuses on how LLMs may enhance market information, enable predictive foresight, and promote tailored, data-driven innovation initiatives while addressing governance and ethical considerations. By reviewing current tools, approaches, and best practices, this study intends to provide a path for enterprises wanting to engage LLMs as strategic partners in fostering sustainable and responsible innovation. [14–18] Figure 1 shows the AI-based innovation management cycle.

Recent research across multiple disciplines has increasingly demonstrated the expanding role of artificial intelligence in enhancing both managerial and biomedical methodologies. In management and organizational studies, AI—particularly Large Language Models—has been widely adopted to improve strategic insight, operational forecasting, and data-driven decision-making across domains such as innovation, HR, marketing, and supply chain management[19–21]. At the same time, biomedical studies have begun incorporating advanced AI techniques, with several works employing Graph Neural Networks and structured frameworks such as the RAIN protocol to identify effective drug combinations and analyze complex gene–protein interactions[22–28]. Notably, these biomedical investigations also integrate LLM-assisted systematic reviews and evidence extraction processes, enabling more accurate synthesis of heterogeneous literature and supporting robust meta-analytic evaluations. Collectively, these developments reflect the broad interdisciplinary impact of modern AI systems and provide a conceptual foundation for understanding how LLM-driven analytical capabilities can support more adaptive, evidence-based, and innovation-oriented strategies.

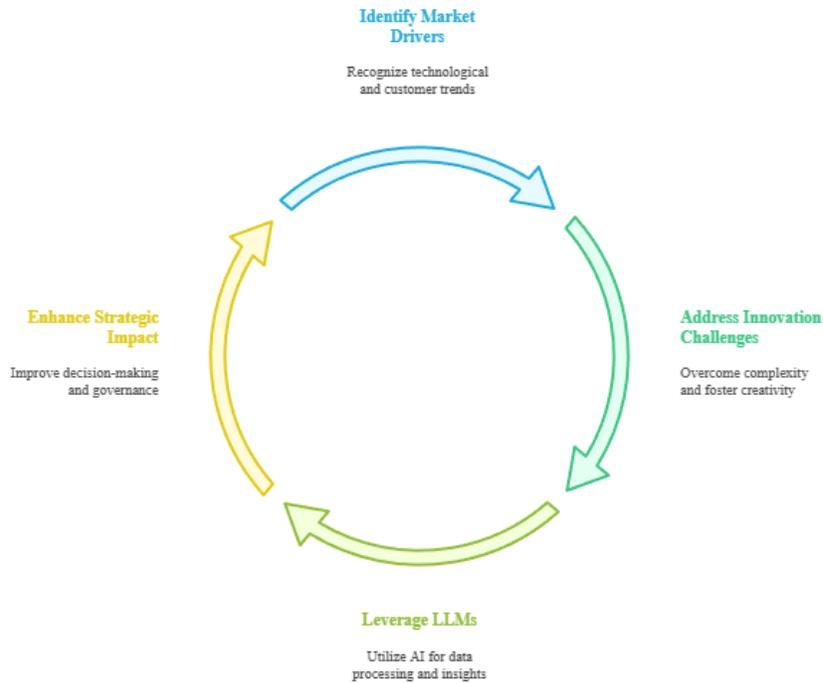

*Figure 1: AI-Driven Innovation Management Cycle*

## 2- Market Intelligence and Predictive Analytics

In today's fast-paced and highly competitive company climate, effective market information and precise predictive analytics are vital for innovation management. Large Language Models (LLMs), with their superior natural language understanding and creation capabilities, provide unparalleled opportunity to better how firms collect, interpret, and act upon market data. LLMs enable the processing of enormous and diverse textual data sources — including social media posts, consumer reviews, news articles, patent filings, and financial reports — to extract actionable insights in real-time. By employing sophisticated natural language processing techniques, these models can anticipate developing trends, detect fluctuations in customer mood, and unearth hidden market signals that traditional analytic tools may overlook. [29–32] Moreover, LLM-driven market data aids competitive research by integrating information on rivals' product launches, marketing activities, and strategic moves. This complete understanding aids firms in anticipating market upheavals and changing innovation strategy proactively. Predictive analytics enabled by LLMs expands beyond descriptive insights, enabling organizations to estimate future market dynamics with enhanced accuracy. By modeling complex linkages within historical and real-time data, LLMs assist in scenario simulation and risk assessment, providing innovation managers with data-driven foresight to optimize resource allocation and prioritize R&D investments. [33,34]

Furthermore, these models help consumer segmentation and personalization tactics by assessing heterogeneous customer data and forecasting behavior trends. This granular intelligence allows for more targeted innovation endeavors, aligning product development with growing market needs. Despite these advantages, integrating LLMs into market intelligence frameworks needs careful consideration of data quality, model interpretability, and ethical concerns. Ensuring openness in AI-driven predictions and reducing potential biases is crucial to preserving stakeholder trust and regulatory compliance. [35–39]

# 3- Adaptive and Personalized Innovation

In today's quickly developing business landscape, innovation is no longer a one-size-fits-all process. Organizations must implement adaptive innovation methods that are tailored to their unique contexts, resources, and market demands. Large Language Models (LLMs) play a vital role in supporting this transformation by delivering scalable, data-driven techniques to adapt and constantly improve innovation activities across all stages of product development. [40–42]

## 3.1 Personalized Innovation Strategies

Personalization in innovation strategy entails adapting ideation, research, and development efforts to the individual needs of different client segments, geographic regions, or business divisions. LLMs support this by evaluating different and frequently unstructured data sources—such as consumer feedback, social media conversations, and competition intelligence—to provide insights that fuel focused innovation roadmaps. By employing enhanced natural language comprehension, LLMs assist decision-makers detect unmet requirements and emerging trends with a level of granularity that was previously unreachable. This enables organizations to select innovation activities that correspond closely with their strategic goals and market realities, boosting the likelihood of commercial success and enduring competitive advantage. [43–47]

## 3.2 Adaptive Learning for Continuous Improvement

Continuous learning and adaptability are critical for maintaining innovation momentum in dynamic situations. LLMs provide adaptive learning by consuming real-time data streams and user feedback, which are then utilized to iteratively update models, recommendations, and tactics. This feedback loop helps firms to evaluate product performance, customer happiness, and market shifts effectively. Consequently, innovation teams can make quick revisions to products, marketing messages, or development objectives, supporting an agile strategy that minimizes time-to-market and mitigates risks associated with static or outdated innovation plans. [48–51]

## 3.3 Customizing Product Development Cycles

Product development cycles must be flexible to accommodate increasing client expectations and technical improvements. LLMs contribute by automating and enhancing many parts of the development lifecycle, from ideation and prototype to testing and launch.

For example, they can produce design options based on current market trends, mimic user feedback using sentiment analysis, and optimize documentation for regulatory compliance. Additionally, by integrating with agile development frameworks, LLMs offer iterative processes, enabling rapid prototyping and continuous delivery that respond directly to stakeholder inputs.

Through this personalization, organizations can balance speed and quality, guaranteeing that new products not only fulfill but anticipate client wants, thereby increasing their innovation pipeline. [52–56] Figure 2 illustrates adaptive and personalized innovation in modern business.

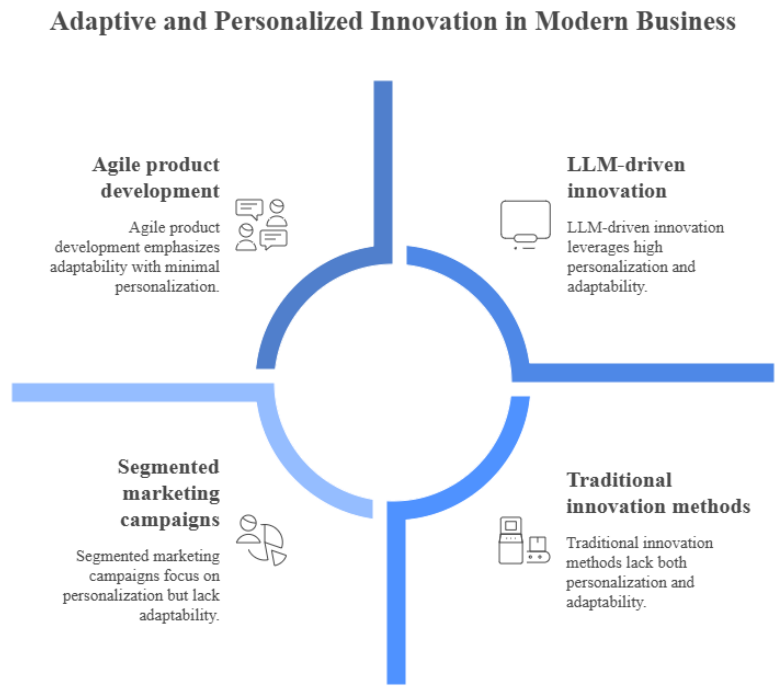

Figure 2: Adaptive and Personalized Innovation in Modern Business

## 4- Collaboration and Data-Driven Decision Support

Product development cycles must be flexible to suit changing client expectations and technical improvements. LLMs contribute by automating and enhancing numerous phases of the development lifecycle, from conceptualization to prototype to testing and launch.

For example, they can provide design options based on current market trends, imitate customer comments using sentiment analysis, and optimize documents for regulatory compliance. Additionally, by interacting with agile development frameworks, LLMs allow iterative procedures, enabling rapid prototyping and continuous delivery that respond directly to stakeholder feedback. [57,58]

Through this customization, enterprises can balance speed and quality, guaranteeing that new products not only fulfill but anticipate client wants, thereby growing their innovation pipeline.

Moreover, LLMs enable multilingual communication capabilities that break down language barriers in increasingly globalized workforces. Real-time translation and localization powered by these models facilitate inclusive participation, ensuring that innovation efforts benefit from diverse perspectives and expertise without being hindered by linguistic limitations. This capability is particularly valuable for multinational corporations managing innovation across diverse markets and cultures. [59–62]

Beyond direct communication facilitation, LLMs function as intelligent intermediates between human decision-makers and huge volumes of organizational and external data. In today's data-rich contexts, deriving actionable insights from unstructured textual information such as emails, reports, market assessments, customer feedback, and social media chatter is a daunting issue. LLMs excel at processing this data, recognizing key trends, and generating natural language explanations that can be queried conversationally. Such data-driven decision support tools democratize access to complex analytics, allowing non-technical stakeholders to interact meaningfully with insights, generate hypotheses, and make evidence-based strategic decisions without having sophisticated data science skills. Integrating LLM-powered tools within collaboration platforms further increases their impact. These solutions encourage dynamic knowledge exchange, enabling quick iteration on ideas, and support agile innovation cycles by providing teams with real-time feedback and contextually relevant information. Importantly, they create a culture of transparency and inclusivity by making knowledge available and minimizing information silos that often inhibit organizational agility. [63–65]

Nevertheless, while the benefits of LLM-enabled collaboration and decision assistance are enormous, businesses must also overcome obstacles related to data protection, model bias, and the accuracy of generated material. Ensuring that LLM outcomes are trustworthy and ethically sound needs ongoing monitoring, human oversight, and integration with rigorous governance systems. In conclusion, the integration of Large Language Models into collaboration and communication processes constitutes a paradigm change in innovation management. By automating regular activities, facilitating multilingual engagement, and enabling data-driven strategic decisions, LLMs empower firms to innovate more effectively, respond promptly to market changes, and sustain competitive advantage in an increasingly complicated global context. [66–68] Figure 3 shows how LLMs accelerate product development through collaboration, communication, and data-driven insights.

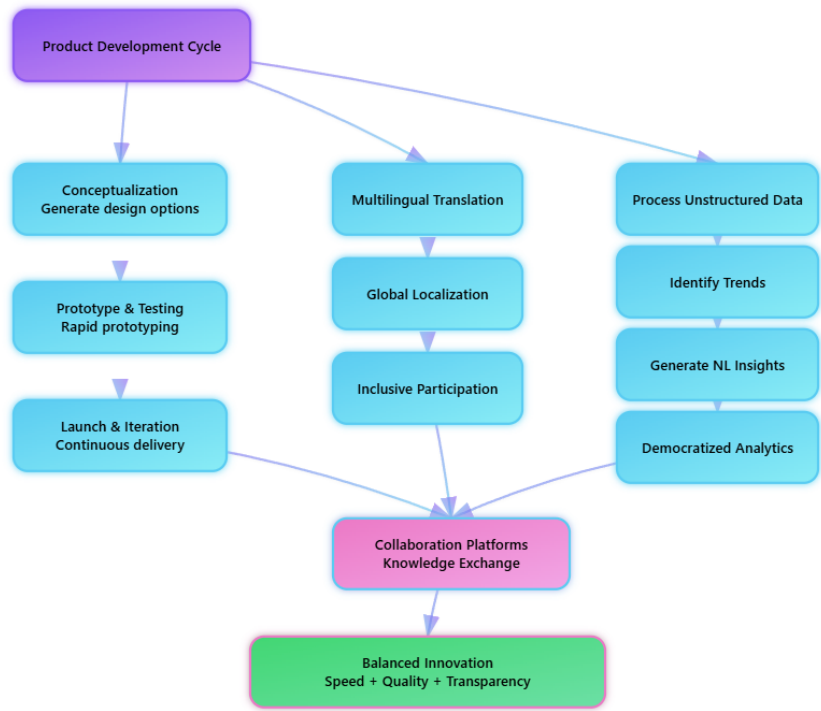

*Figure 3: LLMs driving agile collaboration and product development*

# 5- Ethics, Bias, and Responsible AI Governance

As the adoption of Large Language Models (LLMs) becomes increasingly ubiquitous within innovation management frameworks, the ethical implications of their deployment emerge as key concerns necessitating rigorous attention. The integration of LLMs into strategic decision-making processes, product development, and customer engagement brings forth multiple difficulties relating to fairness, accountability, transparency, and data management. Addressing these difficulties is vital not only to maintain the integrity of innovation ecosystems but also to ensure equitable and socially responsible outcomes. [69–73]

### Bias in LLMs: Origins and Consequences

LLMs are often trained on broad and diverse datasets provided from the internet, literature, and other human-generated content. While this intensive training enables them to capture rich linguistic patterns and global knowledge, it simultaneously exposes them to entrenched cultural prejudices present in the raw data. Such biases might be apparent or subtle, spanning gender, color, ethnicity, cultural prejudices, and more. When LLMs inherit and perpetuate these prejudices, their outputs risk reinforcing existing inequalities, resulting to discriminatory behaviors or exclusionary innovation outcomes.
The ramifications of biased AI outputs are severe, particularly when LLMs influence product recommendations, consumer interactions, employment processes, or policy development within businesses. Biased decisions may affect marginalized populations and undermine the validity of innovation efforts. Therefore, bias mitigation must be a foundational pillar in the development and deployment of LLMs. [74–77]

## Strategies for Bias Detection and Mitigation

LLMs are generally trained on large and diverse datasets provided from the internet, books, and other human-generated content. While this extensive training enables them to catch rich linguistic patterns and global knowledge, it simultaneously exposes them to ingrained cultural prejudices existing in the raw data. Such biases could be visible or subtle, covering gender, color, ethnicity, cultural prejudices, and more. When LLMs inherit and maintain these prejudices, their outputs risk reinforcing existing disparities, resulting to discriminatory behaviors or exclusionary innovative outcomes.

The repercussions of biased AI outputs are serious, particularly when LLMs influence product recommendations, consumer interactions, employment procedures, or policy development within enterprises. Biased decisions may damage underprivileged people and impair the efficacy of innovation efforts. Therefore, bias mitigation must be a foundational pillar in the development and deployment of LLMs. [78–85]

## Responsible AI Governance: Principles and Frameworks

Responsible AI governance is the underlying structure guaranteeing that AI systems function within ethical boundaries and comply with legal rules. Central principles include transparency, accountability, privacy, and human monitoring.

Transparency mandates that AI models and their decision-making logics be interpretable. Users and stakeholders should have access to explanations of how models derive conclusions, enabling examination and trust. Accountability mechanisms establish who is responsible for AI-driven outcomes—whether developers, managers, or end-users—and outline methods for rectifying damages or errors.

Data privacy and security are crucial to effective governance. LLMs often process sensitive proprietary or personal data, necessitating compliance with rules such as the General Data Protection Regulation (GDPR), the California Consumer Privacy Act (CCPA), and other new requirements. Implementing data minimization, encryption, anonymization, and consent management protects individual rights and organizational assets.

Human oversight guarantees that AI augments rather than replaces important human judgment, especially in high-stakes innovation decisions. Hybrid human-AI workflows help balance efficiency advantages with ethical issues. [86–90]

## Risk Management and Scenario Planning

Given the complexity and unpredictability of AI implications, enterprises must implement proactive risk management measures. Scenario planning permits envisioning varied futures—ranging from favorable outcomes to unforeseen consequences such as algorithmic misuse, systemic biases, or market upheavals.

Organizations should build monitoring systems that regularly examine AI behavior post-deployment, recognizing drifts or potential concerns. Contingency plans, including rollback capabilities and human intervention protocols, prepare teams to respond promptly to ethical breaches or failures.

Moreover, cross-sector collaboration, comprising policymakers, academia, public society, and industry stakeholders, helps the establishment of standards and best practices to control LLM use appropriately. [91–96]

### Balancing Innovation and Ethics for Sustainable Growth

Ultimately, the ethical integration of LLMs in innovation management supports not only compliance but also competitive advantage. Companies considered as socially responsible attract talent, customers, and investors, while limiting legal and reputational risks. Embedding justice, transparency, and accountability into AI workflows encourages inclusive innovation ecosystems that provide rewards fairly.
As LLMs evolve, continuing research on bias mitigation, explainability, and governance will be crucial. The future success of AI-powered innovation rests on developing systems that link technology capabilities with social ideals, ensuring that innovation enhances human well-being in an ethical and sustainable manner. [97–99] Figure 4 outlines the principles, risks, and strategies needed to manage AI responsibly and impartially.

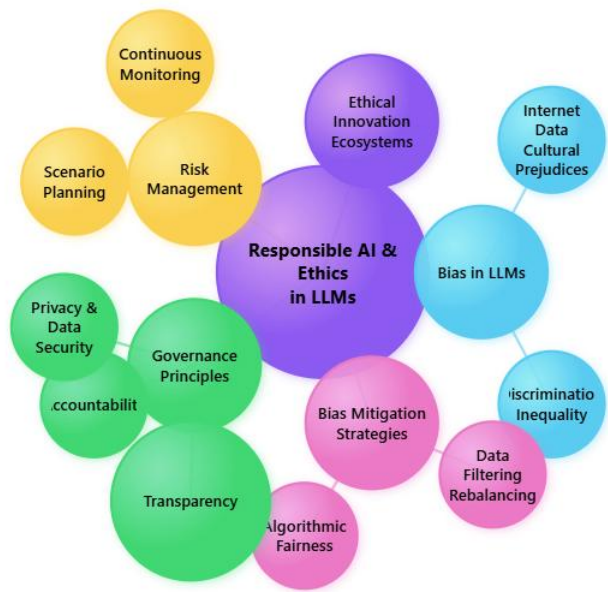

*Figure 4: Responsible AI: ethics, governance, and risk*

## 6- Comparative Review of LLM Models

The environment of Large Language Models (LLMs) has rapidly expanded, offering a varied array of tools and architectures that enterprises can exploit to better their innovation initiatives.

This section gives a comparative examination of famous LLMs, concentrating on their capabilities, scalability, domain flexibility, and applicability for strategic innovation management. Among the most significant models are OpenAI's GPT series, Google's BERT and its derivatives, Meta's LLaMA, and the open-access BLOOM. Each of these models demonstrates unique strengths and trade-offs that impact their efficacy in various innovation environments. [100–105]

GPT series, particularly GPT-4, stands out because to its remarkable generative powers, contextual knowledge, and applicability across applications such as market analysis, trend prediction, and scenario simulation. Its comprehensive training on varied datasets enables nuanced comprehension of industry-specific terminology, making it a strong tool for strategic foresight and decision assistance. However, the proprietary nature and computing demands of GPT-4 can limit accessibility and modification for some businesses. [106–109]

BERT and its derivatives excel in jobs requiring extensive bidirectional context awareness, such as sentiment analysis, customer feedback mining, and document classification. Their pre-training objectives and transformer design enable for successful fine-tuning on domain-specific data, which is vital for customizing innovation strategies to particular market segments or technological areas. Nonetheless, BERT models primarily function as encoders and may require connection with generative modules for entire innovation workflows. [110–117]

Meta's LLaMA delivers efficient training methods and reduced model sizes without sacrificing performance, facilitating deployment in resource-constrained contexts. Its open availability fosters customization and community-driven upgrades, which are helpful for enterprises seeking flexible and transparent AI solutions in their innovation pipelines. [118–121]

BLOOM, as a multilingual open-access LLM, widens the reach of innovation management tools to multiple linguistic and cultural markets. Its ability to digest and synthesize content in various languages promotes global strategy creation and cross-market insights, crucial for multinational organizations. [122–124]

GPT series, notably GPT-4, stands out thanks to its extraordinary generative powers, contextual knowledge, and applicability across applications such as market analysis, trend prediction, and scenario modeling. Its broad training on various datasets enables nuanced interpretation of industry-specific terms, making it a formidable tool for strategic foresight and decision aid. However, the unique nature and processing demands of GPT-4 can limit accessibility and customization for some enterprises. [125–129]

BERT and its derivatives thrive at jobs requiring substantial bidirectional context awareness, such as sentiment analysis, customer feedback mining, and document classification. Their pre-training objectives and transformer design enable for successful fine-tuning on domain-specific data, which is crucial for adapting innovation strategies to particular market segments or technological domains. Nonetheless, BERT models primarily act as encoders and may require connection with generative modules for whole innovation workflows. [130–134]

Meta's LLaMA enables efficient training methods and decreased model sizes without sacrificing performance, facilitating deployment in resource-constrained environments. Its open availability enables customization and community-driven upgrades, which are advantageous for organizations seeking flexible and transparent AI solutions in their innovation pipelines. [135–137]

BLOOM, as a multilingual open-access LLM, increases the accessibility of innovation management tools to many linguistic and cultural markets. Its ability to assimilate and synthesize knowledge in many languages fosters global strategy generation and cross-market insights, vital for multinational firms. [138,139]

The choice of an LLM for innovation strategy depends on various criteria like job specificity, domain requirements, scalability needs, and ethical imperatives. A hybrid approach, integrating the qualities of several models and tailoring them to organizational circumstances, frequently provides the most effective solutions. Future developments are projected to enhance interoperability, eliminate computational obstacles, and encourage more transparent AI-driven innovation environments. Figure 5 compares leading LLMs to identify strengths, limitations, and optimal applications for innovation.

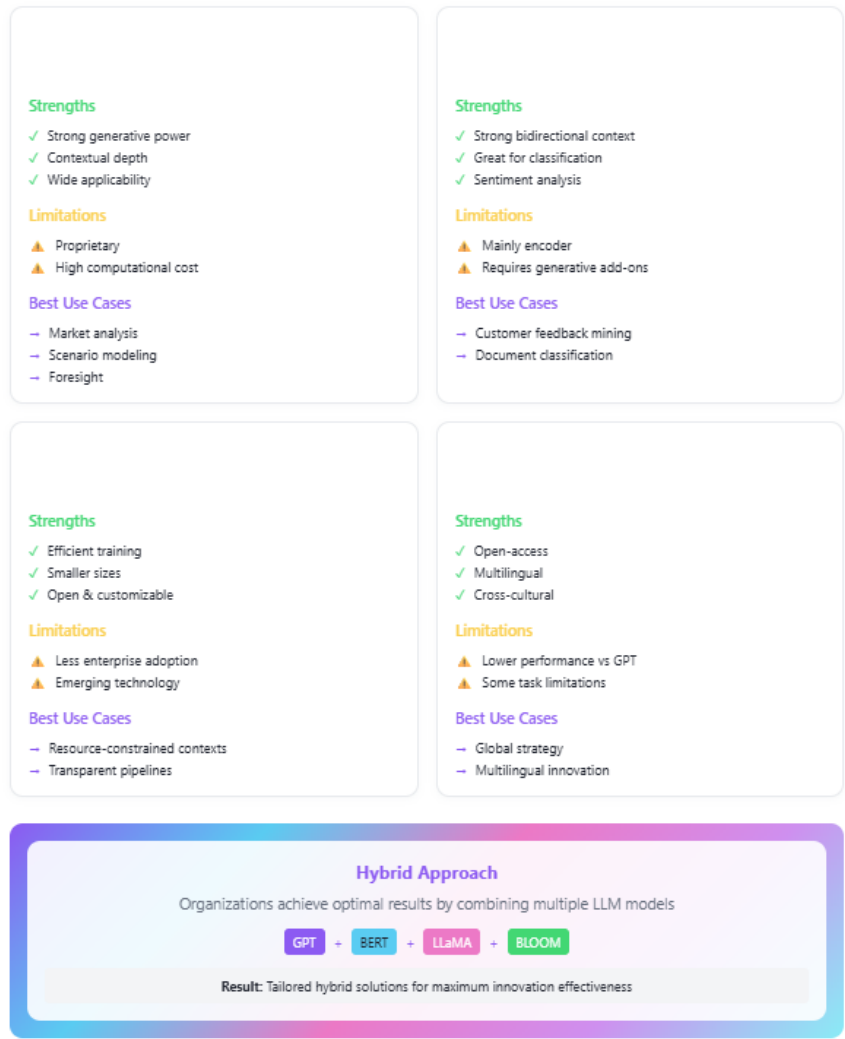

Figure 5: LLM strengths, limits, and best uses

# 7- Future Trends in Sustainable Innovation

The rapid advancement and rising usage of Large Language Models (LLMs) suggest a paradigm shift in the field of innovation management. As these models become more complex, their future evolution will be driven not only by technological developments but also by the compelling need for sustainable and transparent practices. In this setting, several essential routes emerge as focal topics for continuing study and practical implementation. First, the integration of LLMs with multimodal artificial intelligence systems constitutes a key frontier. Moving beyond merely text-based data, future models will combine multiple data modalities—including visual, audio, and sensor inputs—allowing for better contextual comprehension and more complete analysis. Such multimodal frameworks will enable innovation managers to synthesis information from different sources, facilitating nuanced decision-making that accounts for complicated, real-world events. For example, merging product design documents with customer input films or environmental sensor data can discover insights previously unreachable through textual analysis alone. Secondly, the use of privacy-preserving techniques such as federated learning and decentralized model training would be necessary. These approaches allow various businesses or units to collaborate update models without revealing sensitive raw data, therefore adhering to severe data protection rules like GDPR or CCPA. By enabling decentralized knowledge accumulation, federated learning can uncover enormous, previously segregated datasets, boosting the generalizability and robustness of LLMs while guaranteeing user confidentiality and trust. This balance between data utility and privacy will be important to the ethical deployment of AI in innovation environments. [140–142]

Sustainability considerations are increasingly incorporated within the innovation lifecycle, and LLMs are set to play a transformative role here. Future applications will focus on optimizing product creation to minimize environmental footprints, such as through energy-efficient design, waste minimization, and supply chain transparency. By employing LLMs' abilities to analyze complex lifecycle data and regulatory frameworks, organizations may better connect innovation projects with circular economy concepts and international sustainability standards. Moreover, predictive models can anticipate resource scarcity or environmental threats, enabling proactive strategy revisions that assure long-term resilience. Transparency and explainability remain critical challenges as LLMs develop in complexity and relevance. Developing interpretable AI frameworks that explicate model logic will improve confidence among stakeholders, from developers and managers to regulators and consumers. Advances in explainable AI (XAI) approaches, together with open auditing standards and collaborative model governance, will democratize access to AI insights and ensure accountability. Such transparency is not simply a legislative obligation but also a strategic asset, building trust and facilitating human-machine collaboration in innovative processes. [143–152]

Finally, the future roadmap for innovation management emphasizes continual learning and strategic foresight. LLMs will increasingly assist scenario modeling, risk assessment, and dynamic adaptation, helping firms to predict disruptions and respond fast. By incorporating these skills into decision-making workflows, organizations can nurture agility and preserve competitive advantage in uncertain marketplaces. Crucially, this approach focuses on a symbiotic relationship where human judgment and ethical considerations lead AI-driven insights, ensuring that technical advancement translates into equitable and responsible innovation. In conclusion, the trajectory of LLMs in innovation management is distinguished by convergence with multimodal data, privacy-centric learning paradigms, sustainability integration, and transparent governance. Embracing these characteristics will help firms to exploit AI's revolutionary potential while aligning innovation with societal values and environmental sustainability. The route forward involves interdisciplinary collaboration, continual ethical vigilance, and a commitment to sustainable development—an necessity that will define the next generation of innovation leadership. [153–157] Figure 6 highlights emerging trends and sustainable practices that are shaping the future of LLM-based innovation.

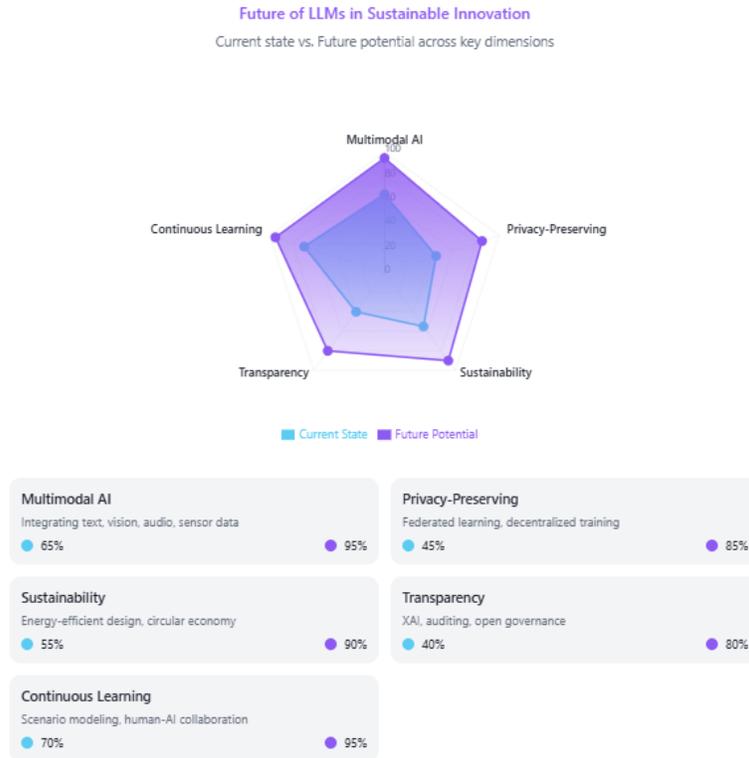

Figure 6: Key trends shaping sustainable AI innovation

## 8- Discussion

The incorporation of Large Language Models (LLMs) into executive-level innovation strategy marks a radical shift in how firms conceive, plan, and execute innovation efforts. As illustrated throughout this study, LLMs allow unmatched capabilities in analyzing large, heterogeneous data streams, delivering nuanced market intelligence, and facilitating adaptive, tailored innovation techniques. However, this integration is neither straightforward nor without substantial strategic and ethical challenges that must be addressed to fully achieve AI's promise as a trusted partner in innovation leadership.

From a strategic standpoint, LLMs complement decision-making by expanding the extent and depth of information available to executives. Their ability to combine massive textual datasets—from patents and scientific literature to real-time market sentiment—facilitates more informed forecasting and scenario planning. This capacity helps leaders to predict disruptions, identify emerging opportunities, and modify innovation portfolios dynamically, eventually boosting organizational agility in volatile markets. Furthermore, LLM-driven insights democratize access to advanced analytics across cross-functional teams, bridging barriers between technical specialists and strategic managers, and building more cohesive and data-informed innovation cultures.

Nonetheless, some significant concerns limit the anticipation for broad LLM adoption at the strategic level. Foremost among these is the ethical governance of AI systems. Biases contained in training data can unwittingly affect strategy priorities, marginalizing specific consumer groups or innovation themes, so compromising inclusivity and social responsibility. Transparency and explainability of AI-driven suggestions remain restricted in many models, complicating accountability and potentially diminishing executive faith in automated insights. Addressing these concerns demands incorporating robust bias detection, ethical review processes, and human-in-the-loop methods that protect crucial human judgment and values within decision workflows. Moreover, the incorporation of LLMs must be properly linked with existing organizational structures and processes. Innovation strategy is intrinsically context-sensitive and shaped by tacit knowledge, cultural nuances, and competitive dynamics that may defy even the most advanced AI models. Executives should consequently consider LLMs as augmentative tools rather than replacements for experiential expertise, ensuring that AI enhances rather than supplants strategic intuition. This demands investment in capability-building, interdisciplinary collaboration, and continual refining of AI systems to meet unique organizational objectives.

Additionally, data governance and privacy considerations create practical restrictions on LLM adoption. Executives must traverse complex regulatory regimes, preserving proprietary information and consumer data while harnessing AI insights. The introduction of privacy-preserving techniques such as federated learning offers intriguing pathways to address these opposing objectives but also brings technical complexity and operational costs. Finally, the fast-paced advancement of LLM technologies necessitates ongoing monitoring and modification. Organizations should build dynamic governance frameworks that enable ongoing review of AI performance, ethical compliance, and strategic alignment. Scenario planning and risk management must consider potential AI-related disruptions, ensuring resilience against model failures, adversarial assaults, or unforeseen outcomes. In summary, while LLMs have considerable promise to transform executive innovation strategy by boosting information gathering, decision support, and adaptive learning, their effective implementation rests on overcoming ethical, organizational, and governance concerns. Embracing a balanced, human-centered strategy that integrates technology capabilities with strategic insight and responsible AI practices will be vital to unlocking lasting competitive advantage in the AI-augmented innovation era. Figure 7 shows the integration of LLMs into the implementation strategy.

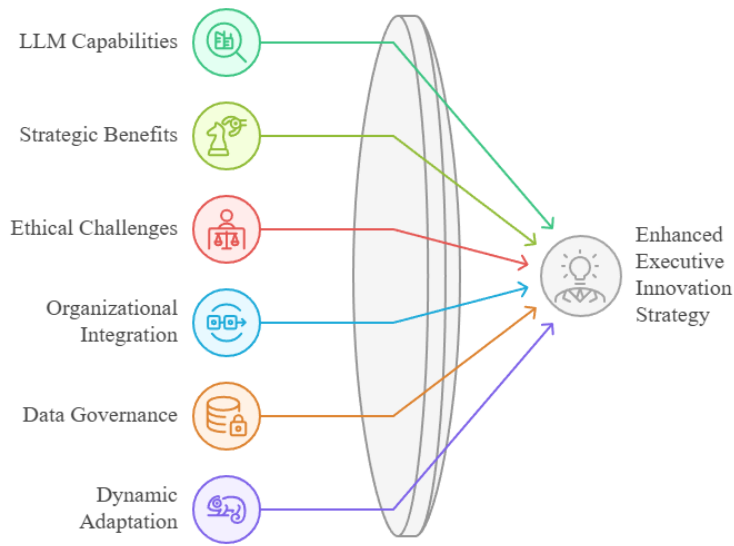

Figure 7: Integrating LLMs into Executive Strategy

# 9- Conclusion

This article has investigated the varied function of Large Language Models (LLMs) in altering innovation management, emphasizing their strategic impact on executive decision-making, market intelligence, adaptive innovation, and collaborative processes. By using the superior capabilities of LLMs to analyze massive and diverse data sources, enterprises may unleash new levels of insight, foresight, and personalization that enable more successful and agile innovation initiatives. However, this technological promise is accompanied with serious ethical and governance issues, including bias mitigation, transparency, privacy protection, and the preservation of human judgment. Responsible deployment of LLMs needs a holistic approach that includes technical safeguards, organizational rules, and continual oversight to ensure AI systems accord with social values and contribute to equitable innovation outcomes.

The comparative review of available LLM tools underlines the need of selecting and modifying models depending on individual organizational situations, balancing performance, scalability, and ethical issues. Looking ahead, the confluence of LLMs with multimodal AI, privacy-preserving approaches, and sustainable innovation principles will further influence the future landscape of innovation management.
Ultimately, organizations that strategically incorporate LLMs within robust ethical frameworks and adaptive governance structures will be best positioned to capitalize on AI-driven opportunities, fostering sustainable growth and maintaining competitive advantage in an increasingly complex and dynamic business environment.